\documentclass[journal]{IEEEtran}
\usepackage{amsmath,amsfonts}
\usepackage{algorithmic}
\usepackage{algorithm}
\usepackage{array}
\usepackage{textcomp}
\usepackage{stfloats}
\usepackage{url}
\usepackage{verbatim}
\usepackage{graphicx}
\usepackage{cite}
\usepackage{bm}
\usepackage{textcomp}
\usepackage{color}
\usepackage{subfigure}
\usepackage[normalem]{ulem}
\newif\ifreview
\reviewfalse  
\ifreview
  \newcommand{\rev}[1]{\textcolor{blue}{#1}}
  \newcommand{\del}[1]{\textcolor{red}{\sout{#1}}}
  \newcommand{\ept}[1]{#1}
\else
  \newcommand{\rev}[1]{#1}
  \newcommand{\del}[1]{}
  \newcommand{\ept}[1]{}
\fi

\usepackage{booktabs} 
\begin{document}

\title{Phase-Conditioned Imitation Learning with Autonomous Failure Recovery for Robust Deformable Object Manipulation}

\author{Dayuan Chen, Kai Tang, Yukuan Zhang, Kazuhiro Kosuge, and Yasuhisa Hirata
\thanks{© 2026 IEEE. Personal use of this material is permitted. Permission from IEEE must be obtained for all other uses, in any current or future media, including reprinting/republishing this material for advertising or promotional purposes, creating new collective works, for resale or redistribution to servers or lists, or reuse of any copyrighted component of this work in other works.}
\thanks{Dayuan Chen, Yukuan Zhang and Yasuhisa Hirata are with the Department of Robotics, Tohoku University, Sendai, Miyagi, Japan (e-mail: chen.dayuan.p3@dc.tohoku.ac.jp; zhang.yukuan.s3@alumni.tohoku.ac.jp; yasuhisa.hirata.b1@tohoku.ac.jp).}
\thanks{Kai Tang and Kazuhiro Kosuge are with the JC STEM Lab of Robotics for Soft Materials, the Department of Electrical and Computer Engineering, Faculty of Engineering, The University of Hong Kong, Hong Kong SAR, China (e-mail: tangkai@eee.hku.hk; kosuge@hku.hk)}
}



\maketitle

\begin{abstract}
\del{This paper presents a force-aware mechatronic framework for the robust manipulation of deformable objects. Unlike open-loop policies, which are prone to failure under occlusion and state aliasing, we propose a closed-loop hierarchical system in which force feedback permeates all components. The architecture consists of three integrated layers: (1) a phase-conditioned imitation learning policy that mitigates aliasing by enforcing deterministic behaviors rather than confused exploration; 
(2) a visual-force phase predictor that monitors task progress and triggers autonomous failure recovery; and (3) a hybrid impedance controller for compliant execution. Furthermore, to ensure dynamic consistency, we introduce a bilateral haptic teleoperation system for collecting contact-rich demonstrations. This explicitly bridges the domain gap between position-based commands and dynamic interaction, which is critical for deformable object manipulation. The system is validated on long-horizon reversible tasks, such as hanging and taking off a t-shirt, demonstrating significant improvements in robustness through active error recovery. The code is available at: https://github.com/leledeyuan00/lerobot/
\newline}
\rev{This paper presents a phase-conditioned, force-aware framework for robust deformable object manipulation.
Standard imitation learning policies such as Action Chunking with Transformers (ACT) rely on a Markovian assumption at inference,
causing state aliasing when visually similar observations require contradictory actions and preventing autonomous recovery from execution failures.
We address this with a closed-loop hierarchical architecture. 
A FiLM-conditioned ACT encoder modulates feature extraction based on the current task phase,
enabling a single unified policy to produce phase-specific behaviors while sharing action dynamics across phases.
A multi-modal phase predictor fusing visual, force, and pose feedback estimates the phase in real time, detecting contact failures that are invisible to vision alone and autonomously triggering recovery trajectories.
The system is completed by a hybrid impedance controller for compliant execution and a haptic teleoperation interface for force-aware data collection.
Ablation studies show that FiLM-based modulation significantly outperforms both unconditioned and token-level conditioned baselines, and t-SNE analysis confirms that FiLM induces well-separated, phase-specific feature representations. Validated on hanging and removing a T-shirt with dual arms, the closed-loop system improves the hanging success rate from 56\% to 87\% through autonomous error recovery.
Code and videos: https://leledeyuan00.github.io/phaser/}

\end{abstract}


\section{Introduction}\label{sec:introduction}

\IEEEPARstart{D}{eformable} object manipulation (DOM), such as manipulating cables and garments, poses challenges in robotics due to the infinite degrees of freedom (DoFs), severe self-occlusion, and complex nonlinear contact dynamics. These characteristics make it extremely difficult to establish accurate analytical models required by traditional control methods. 
Consequently, automation in DOM-related scenarios, ranging from household tasks such as laundry and healthcare assistance to labor-intensive procedures in the garment industry, remains limited. 
To address the challenges of explicit modeling, data-driven approaches have emerged as promising alternatives. 
Reinforcement Learning (RL)~\cite{Yukuan} has demonstrated the capability to discover manipulation policies through interaction. \del{as shown in our previous work}
However, \del{for long-horizon tasks with intricate contact sequences}Imitation Learning (IL) offers \rev{superior sample efficiency for long-horizon tasks by learning directly from human demonstration.}\del{a more sample-efficient approach than designing complex reward functions. It avoids the gap between simulation and reality by leveraging human demonstrations to acquire complex skills directly.}

\begin{figure}[t]
    \centering
    \includegraphics[width=0.95\linewidth]{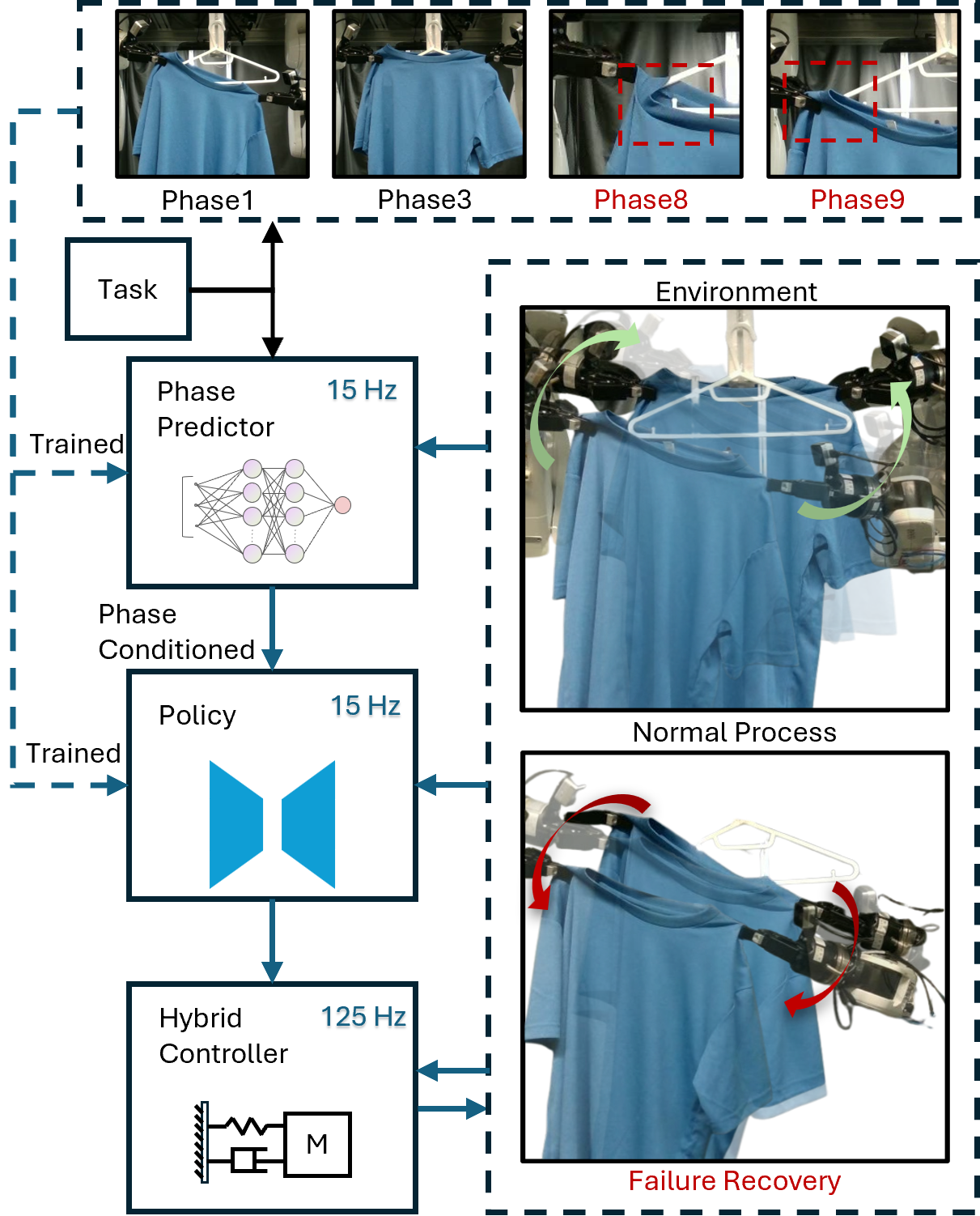}
    \caption{Failure recovery architecture: a phase detector controlling the task process, a phase conditioned policy generating the deterministic behaviors, and a hybrid impedance controller guaranteeing the dynamic adaptivity.}
    \label{Fig: overview}
\end{figure}

Given human-recorded datasets, IL enables robots to acquire complex manipulation skills directly from demonstrations. Prominent methods include Action Chunking with Transformers (ACT)~\cite{ACT} and Diffusion Policy \rev{(DP)}~\cite{DP}. 
However, these approaches typically rely on a Markovian assumption \rev{at inference}, mapping current observations directly to future actions. 
This formulation encounters a critical challenge in long-horizon tasks: state aliasing. 
As illustrated in Fig.~\ref{Fig: overview}, distinct task stages (e.g., the normal insertion phase (Phase~1) and the failure recovery phase (Phase~9)) may exhibit visually similar states yet require opposing action trajectories (e.g., pushing forward vs. retreating). 
Similarly, reversible tasks such as hanging and taking off a garment share similar spatial configurations but require opposite execution directions.
\rev{When trained on data containing such aliased states, the policy learns a mixture of contradictory actions for similar observations, resulting in averaged or oscillatory behavior.
This is not a problem of insufficient data, but a structural ambiguity that requires explicit conditioning to resolve.
}
\del{
Without explicit condition distinctions, such aliasing can confuse the model, leading to averaged or oscillatory behavior that fails to execute the correct sub-task.}

Recent works such as Robot. Transformer (RT-1) \cite{RT-1} and the Diffusion Transformer Policy (DiT) \cite{DiT} employ natural language via \rev{Feature-wise Linear Modulation} (FiLM) \cite{FiLM} layers to \rev{modulate the visual backbone for multi-task}\del{ guide} action generation. 
However, language-based conditioning \del{poses significant limitations for precision engineering: it necessitates}\rev{requires} massive datasets and \rev{introduces} ambiguity, \rev{leading} to unpredictable robot behaviors \rev{that} are unacceptable in contact-rich manipulation. 
For stage-aware execution, Stage-Conditioned Imitation Learning (SCIL)~\cite{SCIL} trains independent per-stage policies switched by a visual observer.
This avoids aliasing within each policy but prevents cross-stage knowledge sharing and scales poorly with the number of stages.
To address these limitations, we inject the phase into the ACT encoder via FiLM, enabling a single unified policy to produce phase-specific behaviors while sharing learned action dynamics across all phases.

Drawing inspiration from human apprenticeship, in which instructors demonstrate not only nominal procedures but also how to correct errors, we explicitly incorporate failure-recovery phases into policy training to enhance robustness at execution time. A closed-loop hierarchical control architecture is proposed for failure recovery. It is controlled by a phase predictor that fuses force, pose, and visual feedback to estimate the current phase in real time, guiding the policy through both nominal and recovery trajectories.
Incorporating force feedback further strengthens the predictor's ability to detect contact failures, such as snagging, that are visually indistinguishable from successful execution due to state aliasing.
The system is further completed by a hybrid impedance controller~\cite{Yukuan-incre} for compliant execution and a bilateral haptic teleoperation interface for collecting force-aware demonstrations.

\del{To translate these high-level policy decisions into safe and compliant physical execution, the control framework must bridge the gap between the neural network and the robot hardware. Since policy inference typically operates at a relatively low frequency (10-15 Hz) and outputs position or velocity commands, direct execution can lead to rigid, jerky motion. To address this, we implemented a hybrid impedance controller based on an incremental impedance formulation \cite{Yukuan-incre}. Operating at a higher control frequency (125 Hz), this bottom-level controller ensures smooth motion and safe physical interaction, serving as the foundation for both data collection and policy deployment. Furthermore, acquiring high-quality demonstrations for contact-rich tasks requires more than just visual coordination; it demands force-aware manipulation. To this end, we developed a bilateral haptic-feedback teleoperation system using the Meta Quest 3 controller, which collects the force response in real time and simultaneously provides vibration feedback to the operator, enabling the human demonstrator to modulate contact forces precisely and confidentially, thereby embedding essential force strategies into the dataset.}

The contribution of this article is summarized as follows:
\begin{itemize}
    \item A \del{novel }phase-conditioned imitation learning \del{framework}\rev{architecture} that \del{introduces}\rev{injects} phase \del{indicators }as \rev{an} explicit dynamics prior\rev{ into the ACT encoder via FiLM}. \del{By decoupling long-horizon garment manipulation into a finite-state machine structure, this effectively resolves}\rev{Unlike token-level conditioning, this mechanism enforces phase-specific feature extraction, resolving} the visual state aliasing \del{problem inherent in deformable object manipulation.}\rev{even when trained on multi-task data with contradictory action trajectories.}
    \item A closed-loop failure recovery mechanism driven by a multi-modal phase predictor. By fusing force, pose, and visual feedback, the system can autonomously detect \rev{visually ambiguous} physical contact failures (e.g., snagging) and trigger recovery policies, \del{significantly enhancing robustness to unmodeled disturbances.}\rev{thereby achieving execution-time robustness.}
    \item An integrated mechatronic system \del{featuring a high-frequency}\rev{with a} hybrid impedance controller and bilateral haptic teleoperation \rev{for compliant execution and force-aware data collection.}\del{interface. This combination ensures compliant physical execution between the robots and the environment and facilitates the collection of high-quality, force-aware demonstration data to capture fine-grained force interaction skills from human demonstrators.}
\end{itemize}

The rest of this article is organized as follows. Section \ref{sec:related work} shows the related works. Section \ref{sec:method} introduces the proposed architecture, which comprises a phase-conditioned ACT policy, a phase prediction network, a hybrid impedance controller, and haptic feedback teleoperation. Experimental results are shown in Section \ref{sec:exp}. Finally, we will give a conclusion and discussion in \ref{sec:conclusion}.

\section{Related Works}\label{sec:related work}
\subsection{Deformable Object Manipulation}
DOM-like garments involve high-dimensional dynamics that are difficult to model analytically. A recent model-based work~\cite{1D2D} employs geometric simplification to accelerate motion planning for 1D and 2D objects, but remains limited to planar tasks. Data-driven approaches address this by explicitly estimating states: the Seam-Informed Strategy (SIS)~\cite{Xuzhao} leverages topological features such as sewing lines, whereas Graph Neural Networks (GNN)-based methods~\cite{Kai, GraphGarment, PanJia} model cloth interactions via graph structures or point clouds. However, these works often suffer from state aliasing when internal contact states are invisible to the camera. Our approach mitigates this by employing 
explicit phase conditioning, which constrains the policy's behavior and prevents divergence caused by self-occlusion and state aliasing.

\subsection{Phase-Conditioned \& \rev{Long-horizon} Imitation Learning}

As discussed in Sec.~\ref{sec:introduction}, generative policies suffer from state aliasing in long-horizon tasks. Language-based conditioning (RT-1~\cite{RT-1}, DiT~\cite{DiT} and stage-conditioned approaches address this from different angles.)
SCIL~\cite{SCIL} \rev{trains independent policies for each stage and switches between them via a gated recurrent unit (GRU)-based visual observer that processes temporal sequences of image features to track task progression. 
However, training separate policies for $N$ stages requires $N$ independent datasets and prevents the sharing of common manipulation primitives across stages.}
Chen et al.~\cite{Deformpam} take a different approach, decomposing long-horizon DOM tasks into action primitives and using preference-based reward learning to select high-quality actions, but do not address intra-task state aliasing or failure recovery.

\begin{figure*}
    \centering
    \includegraphics[width=0.70\linewidth]{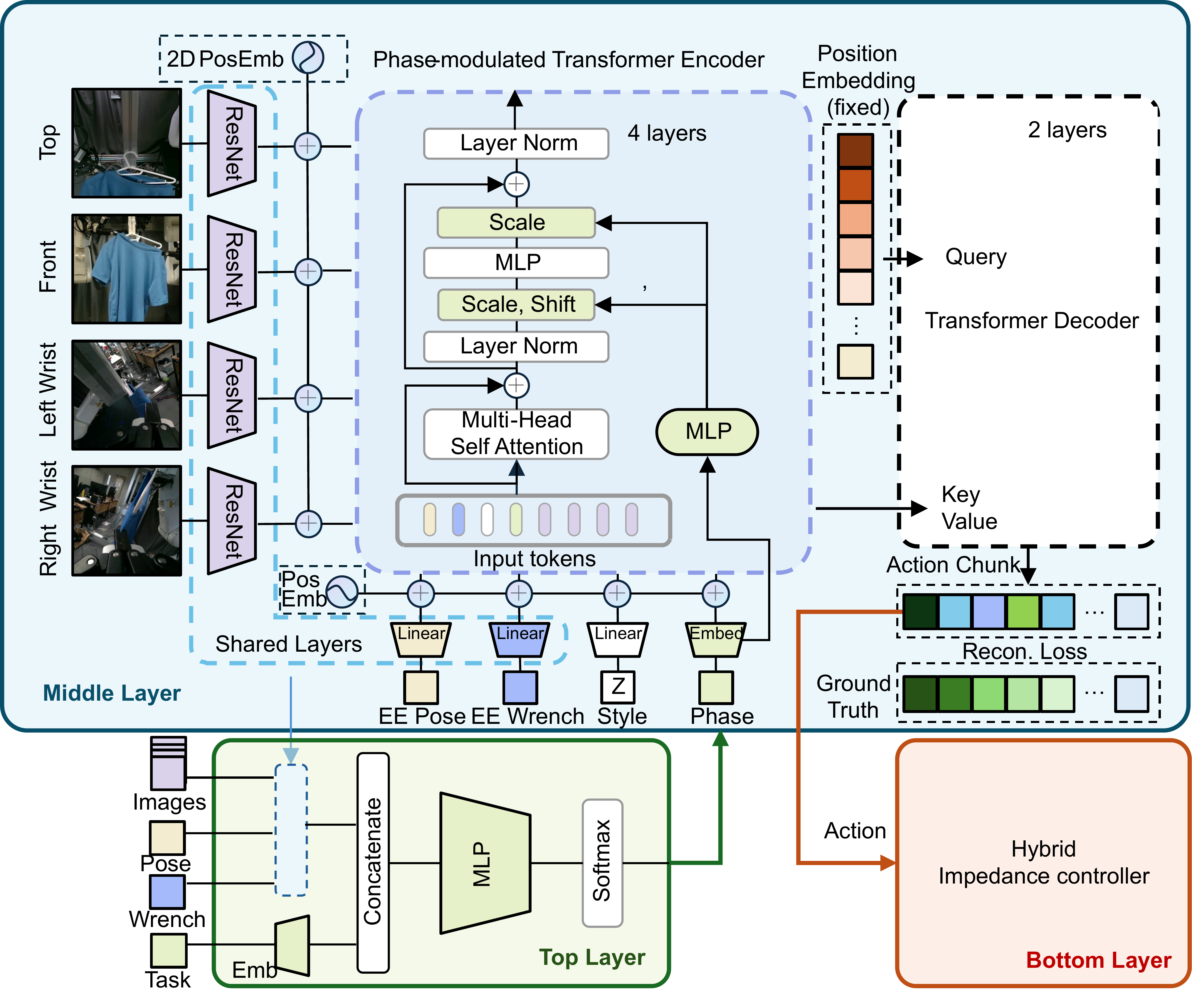}
    \caption{
    Hierarchical Control Architecture. The system is organized into three levels: (Top) The decision layer uses multi-modal feedback to determine the current task phase and monitor for failures; (Middle) The policy layer generates phase-specific actions via the FiLM-conditioned ACT; (Bottom) The interaction layer ensures compliant execution and safety via hybrid impedance control. 
    }
    \label{Fig: model}
\end{figure*}

\del{
However, these state-of-the-art phase-aware methods exhibit critical limitations in 3D contact-rich scenarios. First, the predominantly rely on visual feedback, leaving the system weak at handling occlusion-induced state aliasing, common in 3D garment tasks. Second, their conditioning mechanisms are typically open-loop with respect to physical interactions: SCIL uses phases merely as input tokens without monitoring contact status. At the same time, DeformPAM focuses on planar (2D) tasks, where contact dynamics are simplified.  
Distinct from these primarily visual-based methods, our system incorporates a force-aware phase predictor as a deterministic conditioner. By fusing the force data, we effectively resolve the state ambiguity in 3D contact-rich transitions that visual conditioners often fail to capture.}
\rev{Our method differs from these works in three respects.
First, we train a single unified policy conditioned via FiLM at the transformer encoder level, enabling cross-phase knowledge sharing while producing phase-specific behaviors.
Second, our phase predictor fuses force feedback to detect contact failures invisible to vision-only observers.
Third, the closed-loop integration of predictor and policy enables autonomous failure recovery, which none of the above methods support.}

\subsection{Force-Guided Manipulation \& Recovery}
To resolve visual ambiguity, incorporating force information is essential. 6-axis Force/Torque (F/T) sensing provides a holistic view of the interaction status, making it well-suited for detecting global constraints such as snags or tension limits. Previous works have used F/T feedback \del{for improving}\rev{to improve} assembly~\cite{cartesian_iml} and dual-arm cooperation~\cite{GMM_dualarm} \del{successful rate}\rev{success rates} based on \del{the }Gaussian Mixture Models (GMM).
\rev{Force-Guided Imitation Learning framework with Impedance torque Control (FILIC)}\del{And FILIC}~\cite{force_ACT} \del{using}\rev{incorporates} forces as an input token \del{for}\rev{into} the ACT architecture, which is \del{similar}\rev{closest} to our work\rev{ in terms of multi-modal input design}. However, these frameworks use force only in an open-loop manner, lacking explicit mechanisms to recover from execution failures.
\del{ establishes a closed-loop recovery mechanism. We use force to actively detect failures and trigger specific recovery strategies, ensuring robustness against unmodeled physical disturbances.}

\rev{Beyond force integration, learning recovery behaviors have also been explored. Date Aggregate (Dagger)~\cite{Dagger} and its variants collect on-policy correction data to address distributional shift.
Recovery and Correction (RaC)~\cite{RaC} further standardizes human interventions into recovery-then-correction segments for long-horizon bimanual tasks. However, these methods require human intervention during deployment to trigger recovery and demonstrate it, thereby limiting their autonomy.
In contrast, our framework uses force both as a policy input and as a detection signal in the phase predictor, enabling autonomous closed-loop recovery without human intervention.
}

\section{Methods}\label{sec:method}
\rev{
Standard imitation learning acquires a policy $\pi(a_t | o_t)$ that maps an observation $o_t$ directly to an action $a_t$. 
In long-horizon tasks, however, state aliasing causes distinct phases $p$ to share similar observations while requiring contradictory actions $a_t^*(p)$. 
Without explicit phase information, the policy implicitly learns a mixture $\pi(a_t | o_t) = \sum_p P(p | o_t)\,\pi(a_t | o_t, p)$, whose competing modes produce averaged or oscillatory behavior at aliased states.
}

\rev{
To resolve this, we decompose the policy into a phase-conditioned component $\pi_\theta(a_t | o_t, p)$ and a multi-modal phase predictor $P_\phi(p | o_t)$, 
forming a closed-loop system for autonomous failure detection and recovery. 
As shown in Fig.~\ref{Fig: model}, the architecture comprises three layers: 
a \textit{middle layer} (Sec.~\ref{sec: middle layer}) that realizes $\pi_\theta$ via FiLM-conditioned ACT, producing unambiguous behavior even under multi-task training; 
a \textit{top layer} (Sec.~\ref{sec: top layer}) where $P_\phi$ fuses visual, force, and pose feedback to monitor task progress and trigger recovery phases as a dynamic finite-state machine; and 
a \textit{bottom layer} (Sec.~\ref{sec: bottom layer}) implementing hybrid impedance control for compliant execution. 
A bilateral haptic teleoperation system (Sec.~\ref{sec: teleopration}) is additionally used to collect force-aware demonstrations.
}

\del{We proposed a closed-loop mechatronic system for failure recovery. based on a conditioned imitation learning policy 
It consists of a middle layer: a phase-constrained generative policy that is designed to be not confused by the aliasing state due to multi-task training datasets; A top layer: based on a visual-force phase predictor modelthat generates the phase to control the policy workflow as a dynamic finite state machine; A bottom layer: based on a hybrid impedance controller for making the robot adaptive to environments and keeping the robot and the manipulated object safe. Then, a haptic feedback teleoperation system is proposed to guarantee high-quality position/force-fused datasets. }

\subsection{Middle Layer: Phase Conditioned Hierarchical ACT}
\label{sec: middle layer}

\rev{
A key design decision is how the phase condition $p$ modulates the policy.
A straightforward approach is to inject $p$ as an additional input token to the transformer encoder.
However, since the encoder processes over 1200 visual tokens from four camera views,
a single phase token is easily diluted during self-attention and exerts minimal influence on feature extraction, which we confirm empirically in Sec.~\ref{sec: t-sne}.}

\rev{
Instead, we adopt FiLM, which applies affine transformations directly to the layer normalization and MLP outputs within every encoder layer.
This multiplicative mechanism ensures that the phase condition modulates every intermediate representation, making it impossible for the encoder to ignore.}

\rev{
We apply FiLM to the encoder rather than the decoder for two reasons.
First, state aliasing originates at the perceptual level.
And visually similar inputs must be mapped to distinct feature representations prior to action generation.
Second, keeping the decoder phase-agnostic encourages it to learn shared low-level action dynamics across phases and tasks, such as reaching, inserting, and retreating.
This improves data efficiency when demonstration data are limited.
}

\begin{figure}[t]
    \centering
    \includegraphics[width=0.95\linewidth]{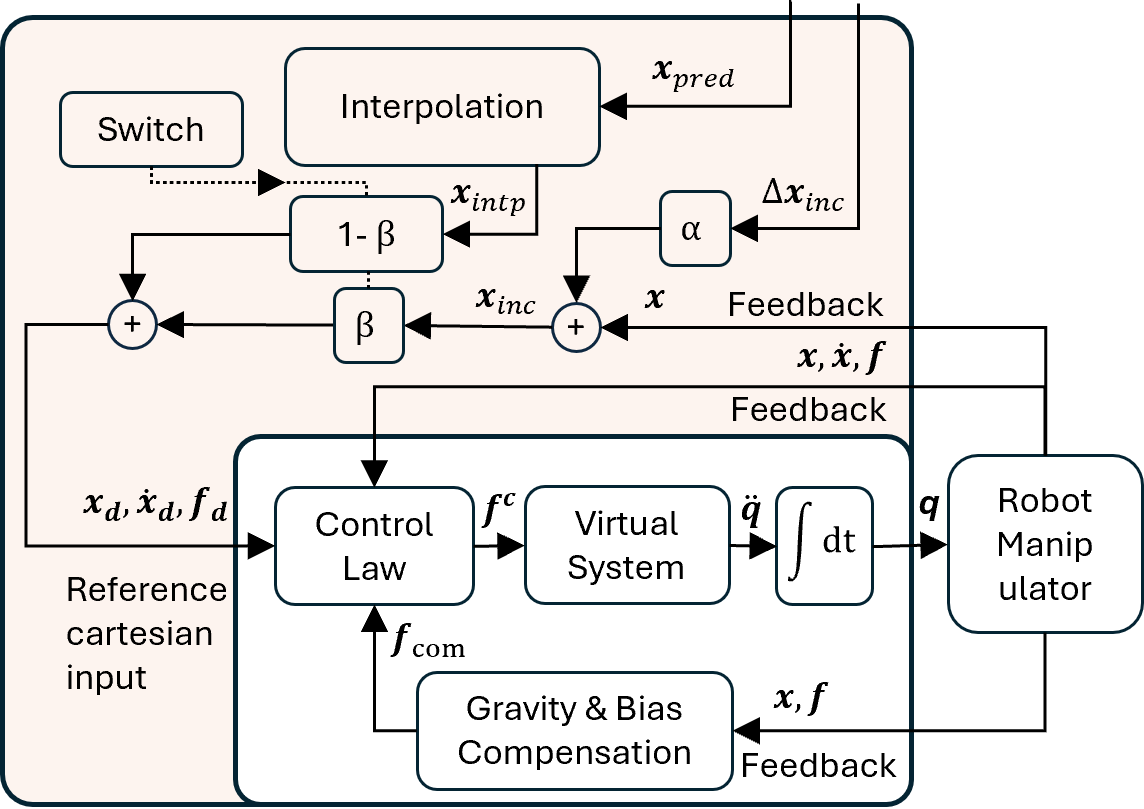}
    \caption{Hybrid impedance controller.}
    \label{Fig: control diagram}
\end{figure}

\del{
This model uses the ACT \cite{ACT} as a backbone. We insert a FiLM \cite{FiLM} layer, similar to DP \cite{DP}, into Encoder layers rather than Decoder layers, as DP did. The idea is to treat the decoder as a skill generator. These phases and tasks, such as hanging and taking off a T-shirt, should exhibit similar dynamics.}
\del{
Sharing the decoder can encourage the model to learn the common low-level action-generation dynamics across these phases and tasks. Additionally, the encoder is used to extract features, and the FiLM layer is used to control this process by modulating internal layers of the encoder to achieve a distinct perceptual attention for task-relevant features.}


\del{
The model takes 4 RGB images, each image with $D_t \in \mathbb{R}^{H \times W \times 3}$, then go through a shared ResNet-18 backbone into a feature map of size $\mathbb{R}^{H^{'} \times W^{'} \times d_\text{model}}$, where $d_\text{model}$ is predefined model hidden dimension using for tokens and transformer layers. The feature then is flattened along the spatial dimension to obtain $(H^{'} W^{'}) \times d_\text{model}$ tokens. 
To preserve the spatial information, a 2D sinusoidal positional embedding is added to the feature sequence. 
We also use the positions and orientations of the dual robots' end-effectors (EE) and the two wrenches of the dual robots as inputs. For rotation in training and inference, we use 6D \cite{Rot6D} instead of a quaternion representation for achieving a stable prediction and execution. The input of robot state is of size $\mathbb{R}^{20}$, which includes 6 for positions and 12 for orientations. At the same time, the wrench is of size $\mathbb{R} ^{12}$. Images, positions, and wrenches are normalized (without rotation) and then used as input to the model. 
We use three linear models that map the pose, wrench, and a style variable to $d_\text{model}$. 
The phase is mapped by an embedded layer from an integer to $d_\text{model}$. 
A 1d positional embedding from the embedded layer is added to these input sequences. Finally, these input tokens are stacked on $\mathbb{R}^{(4H^{'}W^{'}+4)\times d_\text{model}}$ into the to the self-attention transformer encoder.
}

\rev{The model processes four RGB images through a shared ResNet-18 backbone.
The resulting feature maps are flattened into $(4H^{'} W^{'}) \times d_\text{model}$ tokens,
augmented with 2D sinusoidal positional embeddings.
The dual-arm end-effector poses ($\mathbb{R}^{20}$, using 6D rotation~\cite{Rot6D}) and wrenches ($\mathbb{R} ^{12}$) are each projected to $d_\text{model}$ via linear layers. 
The phase label is mapped through a learned embedding.
All tokens are concatenated into a $(4H'W'+4) \times d_\text{model}$ sequence and fed into the transformer encoder, where FiLM modulates each layer as described above.
}

\del{
Where the style variable $z$ is output by the transformer-based Conditional Variational Autoencoder (CVAE). It takes as input the EE pose, wrenches, embedded phase, a target action sequence of length $N_h$, and a learned prepend "[CLS]" token, forming a ($N_h + 4$) sequence. Only the first output of the CAVE encoder is used as the "[CLS]" feature, which is queried to generate the mean and variance of $z$, which is sampled for training and set to the prior mean (zero) at inference for deterministic decoding. Then $z$ is projected to $d_\text{model}$ into the ACT encoder as mentioned above.}

\rev{A Conditional Variational Autoencoder (CVAE) encoder produces the style variable $z$.
During training, it encodes the ground-truth action sequence to learn a latent distribution.
In inference, $z$ is set to the prior mean for deterministic decoding.
}
\rev{
The decoder follows the standard ACT architecture~\cite{ACT}.
Fixed positional queries attend to encoder outputs via cross-attention.
The output is projected to action chunks of horizon $N_h$:
$a_t = [a_{t,L}^{\top}, a_{t,R}^{\top}]^{\top} \in \mathbb{R}^{20}$, 
where each arm's action comprises position ($\mathbb{R}^3$), 6D orientation ($\mathbb{R}^{6}$),
and gripper command ($\mathbb{R}^{1}$).
}

\del{
The FiLM layer is a multi-layer perceptron (MLP) that uses the embedded phase as input and generates a $3\times d_\text{model}$ output, which is divided into $\alpha,\beta,\gamma$ for each $d_\text{model}$. The $\alpha$ is a scale operation after MLP in ACT encoder, and the $\ beta$, $ \gamma$ are scale and shift operations after layer normalizer (LN) of multi-head self attention (MSA). FiLM has strong performance in modulating features; after this, the encoder's behavior can be firmly controlled by phase, without confusion in aliasing states, and the policy will focus on generating a specific, well-defined behavior.}

\del{
The decoder is the standard ACT decoder architecture, which uses an embedded action-position sequence of size $N_h \times d_\text{model}$ as the query and outputs from the encoder as the key and value via a cross-attention mechanism to generate a sequence of predicted action chunks of size $N_h \times d_\text{model}$ down-projected by an MLP to $N_h \times 20$, representing manipulator actions $a_t = [a_{t,L}^{\top}, a_{t,R}^{\top}]^{\top} \in \mathbb{R}^{20}$. Each $a_{t,*}^\top$ consists of the manipulator's position $\mathbb{R}^3$ and 6D orientation $\mathbb{R}^{12}$, as well as the gripper's command $\mathbb{R}^{1}$.}

\subsection{Top Layer: Phase Prediction and Failure Recovery}
\label{sec: top layer}

\begin{figure}[t]
    \centering
    \includegraphics[width=0.64\linewidth]{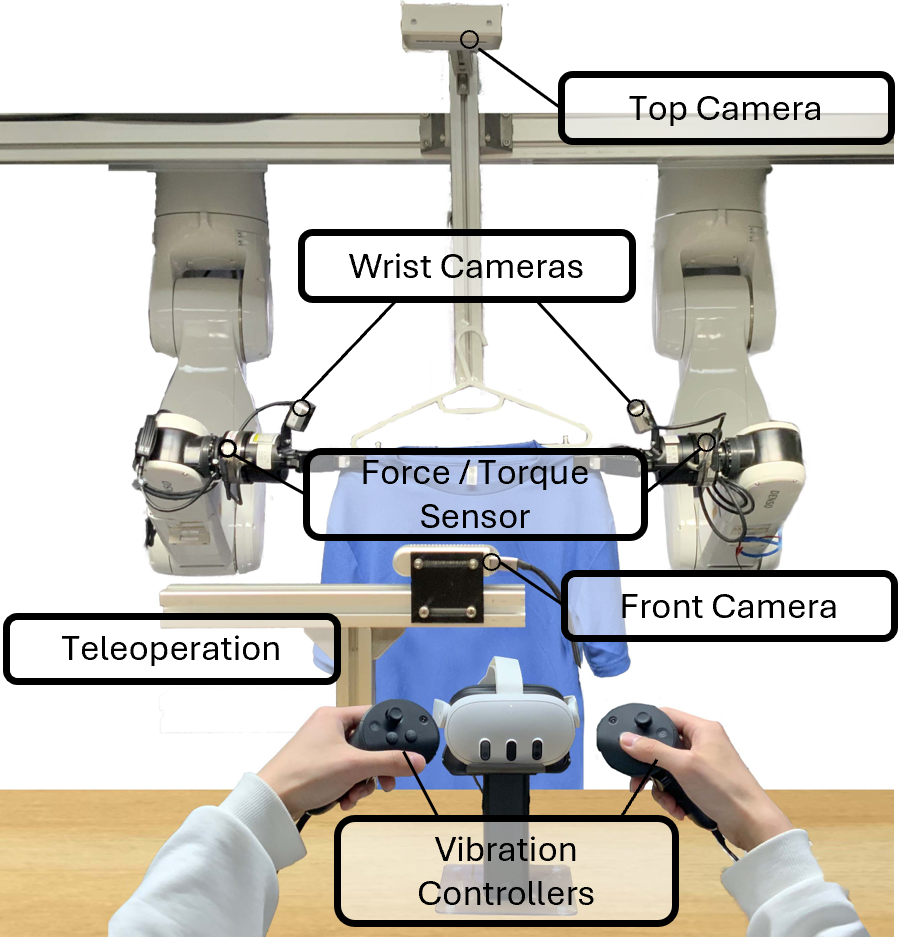}
    \caption{The system overview with haptic feedback teleoperation.}
    \label{Fig: system}
\end{figure}

\del{
After the model can generate deterministic behavior controlled by phase, the next step is to develop a stable phase generator to maintain the workflow as a robust, stable system.
The difficulty of modeling a deformable object made it challenging to design state perception explicitly by manual design. 
Therefore, we use a smaller, shared policy-learned backbone feature extractor to predict the phase used in the imitation policy, thereby sharing as many of the trained capabilities as possible, as shown in Fig. \ref{Fig: model}.
}

\rev{
The phase-conditioned policy generates deterministic behavior given a specific phase.
In deployment, however, the phase must be estimated in real time.
By feeding the predictor’s output back into the policy to condition it, 
we construct a closed-loop mechanism: the phase predictor monitors task progress, detects failures, and outputs the corresponding recovery phase to redirect the policy.
Once recovery succeeds, the predictor reverts to the normal phase and resumes nominal execution.
}

\rev{
The accuracy of failure detection is critical to this loop.
In contact-rich DOM, the most common failure modes, such as snagging and jamming, are visually ambiguous.
Occlusion can make a failed insertion appear nearly identical to a successful one in the camera view.
However, these failures produce distinct signatures in the force domain, such as spikes or sustained abnormal forces.
Involving force feedback in the phase predictor, therefore compensates for the inherent limitation of vision-only detection.
}

\del{
The feature map from the image backbone will be processed by global average pooling (GAP) to continue downsampling from $H^{'} \times W^{'}\times d_\text{model}$ to $1 \times d_\text{model}$.
Moreover, since the visual states of `starting to hang` and `finishing takeoff` can be spatially similar, injecting the Task ID acts as a constraint before constraining the phase prediction search space, after an embedded layer mapping to $1 \times d_\text{model}$}

\del{
Finally, we concatenate all projected features along the feature dimension to obtain a $1 \times 7d_\text{model}$. An MLP projects the input features to the number of phases, and after a softmax, the phase is predicted.}

\rev{
The phase predictor shares the frozen image backbone and pose/wrench projection layers from the trained ACT policy.
Each camera's feature map is reduced to $1 \times d_\text{model}$ via global average pooling (GAP).
The dual-arm poses and wrenches are each projected to $1 \times d_\text{model}$ as in the policy network.
A Task ID embedding is additionally injected to constrain the prediction search space, since phases such as `starting to hang` and `finishing takeoff` share similar visual configurations.
All tokens are concatenated into a $1 \times 7d_\text{model}$ feature vector, which an MLP maps to phase logits followed by softmax.
}

\del{
We found that the behavior of the two transited phases might exhibit markedly different characteristics, so it is better to reset the ACT policy temporal ensembler, which is used to smooth action chunks \cite{ACT}. We also use a 1-second low-pass filter to stabilize and smooth the phase transition.}

\del{
During training, we use all datasets for conditioned-ACT training and record several static datasets to strengthen the performance of phase (nominal and failure) detection. This means the architecture is flexible and can be designed to accept more diverse conditioned inputs without affecting the policy. }

\rev{
During phase transitions, the behaviors of adjacent phases may differ substantially.
We therefore reset the ACT temporal ensembler~\cite{ACT} at each transition and apply a 1-second low-pass filter to suppress spurious phase switches.
For training, we use the full demonstration dataset for the conditioned-ACT policy and additionally collect static-scene samples to improve detection of both nominal and failure phases.
}


\subsection{Bottom Layer: Hybrid Impedance Controller}
\label{sec: bottom layer}

\del{
The policy predicts a target EE pose $\bm{x}_{pred}$ at 15 Hz based on the EE pose observed $\bm{x}_{obs}$ at the policy inference time.
To improve control performance and ensure system safety, rather than directly using the policy output, we use it in two ways, which we call the hybrid impedance controller, as shown in Fig. \ref{Fig: control diagram}. The one directly uses the low-frequency $\bm{x}_{pred}$ to interpolate a high-frequency smooth trajectory $\bm{x}_{intp}$. 
The other is that we compute an incremental command $\Delta \bm{x}_{icnc} = \bm{x}_{pred} - \bm{x}_{obs}$, which is added to the current EE pose $\bm{x}$ to form the final control target $\bm{x}_{inc}$. Then, using two parameters $\alpha$ and $\beta$ combined with $\bm{x}_{intp}$ to a flexible adjusting system response.
This formulation induces behavior similar to implicit force control, where pose commands are executed through an impedance controller. We refer to this scheme as an Incremental Impedance Controller (Inc-IC) \cite{Yukuan-incre}. }

The policy predicts a target EE pose $\bm{x}_{pred}$ at 15\,Hz based on the observed pose $\bm{x}_{obs}$.
Rather than executing $\bm{x}_{pred}$ directly, we employ a hybrid impedance controller (Fig.~\ref{Fig: control diagram}) that combines two command streams.
The first interpolates $\bm{x}_{pred}$ into a high-frequency smooth trajectory $\bm{x}_{intp}$.
The second computes an incremental command $\Delta \bm{x}_{icnc} = \bm{x}_{pred} - \bm{x}_{obs}$ and adds it to the current pose $\bm{x}$ to form $\bm{x}_{inc}$. 
Two parameters $\alpha$ and $\beta$ blend these streams, inducing behavior similar to implicit force control.
We refer to this scheme as an Incremental Impedance Controller (Inc-IC)~\cite{Yukuan-incre}.

\del{
In this work, we switch control modes based on the gripper state by controlling the $\beta$.
When the gripper closed, we assume the robot is in contact with the manipulated object and enable a force-oriented control mode to facilitate interaction.
When the gripper opened, the controller switched to pure impedance control to maintain system stability and prevent drift caused by fluctuating contact forces.}

\del{
Both controller commands will be sent to an impedance controller with gravity compensation \cite{GravityCompensation}, which maintains the force response as the robot's orientation changes. 
Moreover, this impedance controller is designed for position/velocity interface robots, especially for industrial robots \cite{FDCC}. 
It uses a forward virtual-dynamics kinematics, which uses $H_q J^\top$ instead of just $J^{-1}$, mapping the Cartesian space commands to joint space, thereby maintaining safe operation near singularities and limit positions, where $J$ is the Jacobian matrix updated in real-time and $H$ denotes the mechanism's positive definite joint space inertial matrix.}

\rev{
The parameter $\beta$ switches the control mode based on the gripper state.
When the gripper is closed, the robot is assumed to be in contact with the object, and a force-oriented mode is enabled to facilitate compliant interaction.
When the gripper is open, the controller reverts to pure impedance control to maintain system stability and prevent drift from fluctuating contact forces.
Both command streams are executed through an impedance controller with gravity compensation~\cite{GravityCompensation}, designed for position/velocity interface industrial robots~\cite{FDCC}.
}

\del{
The impedance controller is defined by Eq. \ref{Eq: Impedance formation}.
}

\ept{
\begin{equation}
\del{
    \begin{aligned}
    & \Delta\bm{x}=\bm{x}_d-\bm{x}, \\
    & \bm{f}_{imp} = \bm{M} \Delta\bm{\ddot{x}}+\bm{D} \Delta\bm{
    \dot{x}}+\bm{K} \Delta\bm{x},
    \end{aligned}
    \label{Eq: Impedance formation}
}
\end{equation}
}
\del{
where $\bm x_d$ represents the desired position and orientation, $\bm x$ is the current state of the system, $\bm M$, $\bm D$, and $\bm K$ are the designed mass, damping, and stiffness matrices, respectively.
}

\del{
By combining the real-time force/torque data from the sensor with the compensation from the previous section, we can calculate the desired force for the robotic system.
}

\ept{
\begin{equation}
\del{
    \bm{f}^n = \bm{f}_{imp} + \left[ \bm{f}_d-\left(\bm R_e^b \bm{f}_s-\bm{f}_{com}\right) \right],
    \label{Eq: force equation}
    }
\end{equation}
}
\del{
where $\bm{f}_d$ represents the desired force and torque while $\bm{f}_{com}$ is the compensated force/torque component \cite{GravityCompensation}.
}

\del{
A Proportional-Derivative (PD) controller is used to ensure that the system achieves the desired force, as shown in Eq. \ref{Eq: PD controller}.
}
\ept{
\begin{equation}
\del{
    \bm{{f}}^c=\bm K_P \bm{f}^n+\bm K_D\dot{\bm{f}}^n,
    \label{Eq: PD controller}
}
\end{equation}
}
\del{
where $\bm K_P$ and $\bm K_D$ are the proportional and derivative gains. $\Delta t$ is the control period of the system.
}

\del{
Using the manipulator dynamics, the desired force and torque at the end-effector can be mapped to the corresponding torques at each joint as shown in Eq. \ref{Eq: robot dynamics}.}
\ept{
\begin{equation}
\del{
    \bm \tau_{(q)} = \bm J^T \bm{f}^c.
    \label{Eq: robot dynamics}
}
\end{equation}
}

\del{
To translate the joint torques into joint accelerations, an estimated dynamics mapping matrix $\bm{\hat{H}}_{(q)}$ is derived from the current state using the kinematics dynamics library (KDL). This mapping is expressed in Eq. \ref{Eq: Virtual dynamics kinematics}. We refer the reader to \cite{FDCC} for more details on this aspect.}
\ept{
\begin{equation}
\del{
    \bm{\ddot{q}}= \bm{\hat{{H}}}^{-1}_{(q)} \bm\tau_{(q)}.
    \label{Eq: Virtual dynamics kinematics}
    }
\end{equation}
}

\del{
By integrating this acceleration, the desired joint position commands are obtained, as described in Eq. \ref{Eq: Integration for joint command}. }
\ept{
\begin{equation}
\del{
    \left\{\begin{aligned}
    & \bm q_t=\bm q_{t-1}+ \alpha (\bm{\dot{q}}_{t-1} \Delta \hat{t}), \\
    & \bm{\dot{q}}_t=\bm{\dot{q}}_{t-1}+ \alpha (\bm{\ddot{q}}_{t-1}\Delta \hat{t}),
    \end{aligned}\right.
    \label{Eq: Integration for joint command}
    }
\end{equation}}
\del{
where $\alpha$ is the damping parameter, which is applied during integration to prevent oscillations, $\Delta \hat{t}$ is the virtual control update period, which does not necessarily need to match the real control period.}

\begin{figure*}
    \centering
    \includegraphics[width=0.95\linewidth]{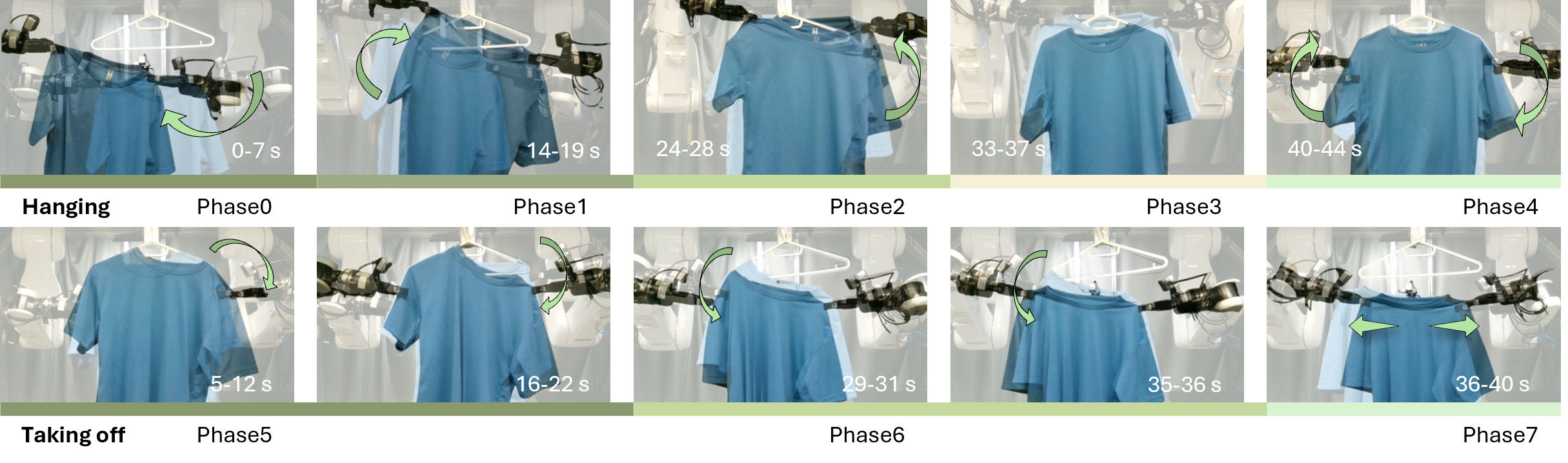}
    \caption{ Upper: Sequence of the nominal hanging task. The process is decomposed into: (Phase 0-1) Collaborative insertion of the left collar; (Phase 2) Right collar insertion; (Phase 3) Lifting and centering; and (Phase 4) Final alignment and release.
    Bottom: Sequence of nominal taking-off task. The robot sequentially grasps and drags the right sleeve (Phase 5) and the left sleeve (Phase 6) to remove the garment, then returns to the home position (Phase 7).
    }
    \label{Fig: exp-Nominal_tasks}
\end{figure*}

\subsection{Haptic Feedback Teleoperation System}
\label{sec: teleopration}
\del{
To ensure the safety of the dataset collection and maintain consistency throughout the pipeline, we also employ an impedance controller for the teleoperation dataset collection process. Furthermore, we use vibration as bilateral haptic feedback to cue the human operator on contact and manipulation force information. 
Otherwise, due to state aliasing, even the human operator cannot clearly and rapidly verify the contacting states solely from visual information.
These ensure that the collected dataset implicitly encodes valid force-compliance strategies and guarantees its quality.}


\rev{
The teleoperation system reuses the impedance controller from Sec.~\ref{sec: bottom layer} to ensure compliant interaction during data collection.
Bilateral haptic feedback is provided via the Meta Quest~Pro controller: real-time force sensor readings are mapped to vibration with coupled amplitude and frequency~\cite{OrbikEbert2021OculusReader}, enabling the operator to perceive contact states that are visually ambiguous due to occlusion.
This ensures the collected demonstrations implicitly encode force-compliance strategies critical for contact-rich manipulation.
}

\del{
During the operation, we found operators were not sensitive to whether only the set vibration amplitude on the controller was mapped from the force sensor information. We then designed a real-time, simultaneous method for changing amplitude and frequency to improve the operator's experience based on \cite{OrbikEbert2021OculusReader}. We set a dynamic 500 ms window as the vibration period, with each vibration lasting around 50 ms. As the force increases, the amplitude and frequency during the dynamic period increase. For example, when the force is near 0, it will vibrate only once during the 500 ms with a small amplitude. Until the force is near 1, the vibration tends to continue during the window.}



\del{
This package can be easily integrated on different operating systems, and published on \url{https://github.com/leledeyuan00/oculus_reader_haptics}, which was developed based on \cite{OrbikEbert2021OculusReader}.}

\section{Experiment}\label{sec:exp}
In this section, we present two tasks: hanging and taking off a t-shirt (Fig. \ref{Fig: exp-Nominal_tasks}) to demonstrate the validity of the proposed phase-constrained policy and the closed-loop system.



\begin{figure*}
    \centering
    \includegraphics[width=0.95\linewidth]{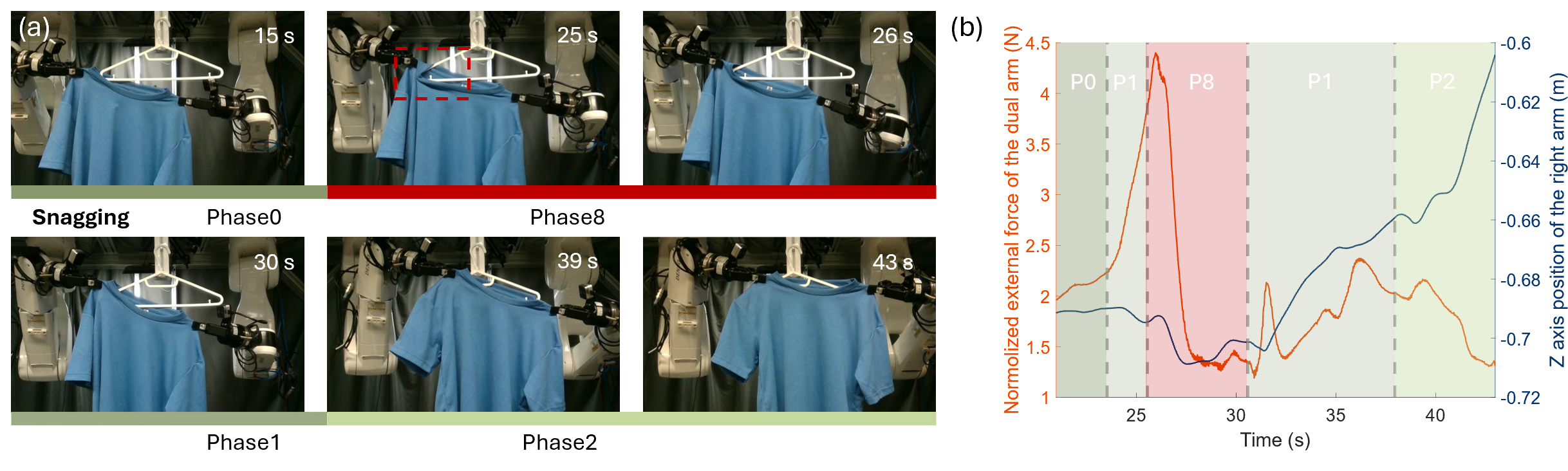}
    \caption{
    Recovery from insertion snagging (Phase 8). (a) Visual serials: Upon detecting a snag, the policy triggers a shaking motion to insert the collar. (b) Data profile: The left curve shows a spike in summed external force (dual-arm) indicating snag, while the right curve tracks the z-axis motion during the recovery maneuver.
    }
    \label{Fig: exp-failure phase8}
\end{figure*}

\begin{figure*}
    \centering
    \includegraphics[width=0.95\linewidth]{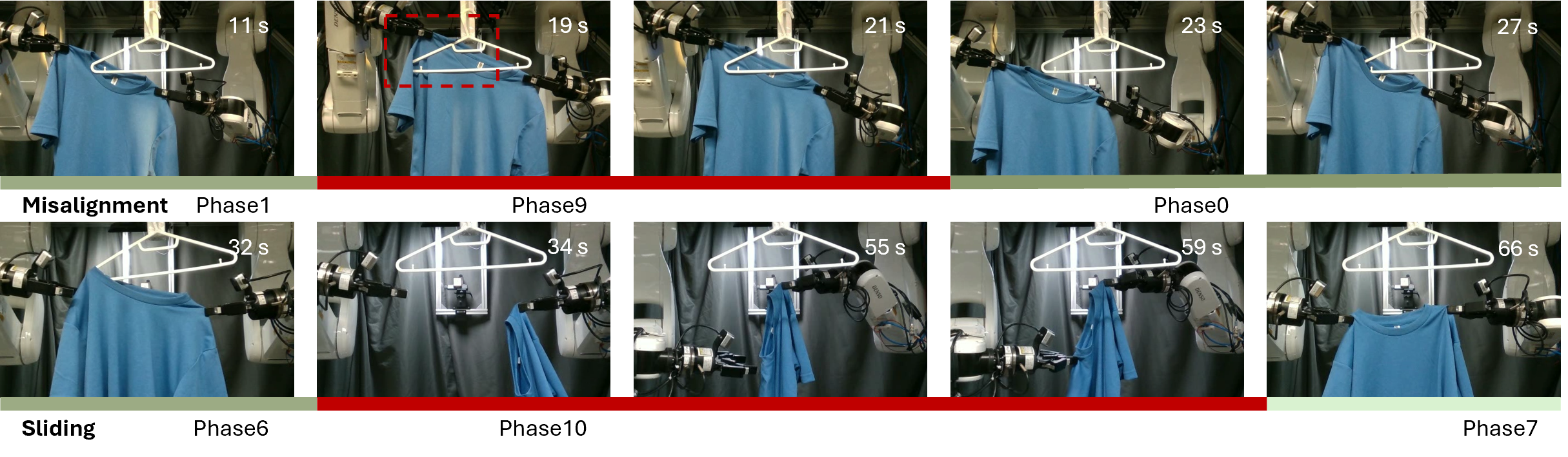}
    \caption{
    Upper: Recovery from gross misalignment (Phase 9). When the garment misses the hanger entirely (front/back), the policy executes a retreat maneuver to reset the system to Phase 0 for a retry.
    Bottom: Recovery from sliding (Phase 10). If the t-shirt prematurely slides off the hanger while taking off, the system transitions to a dynamic grasping-in-the-air strategy to catch the falling garment.
    }
    \label{Fig: exp-failure phase9/10}
\end{figure*}

\subsection{Hardware Description}
The manipulators are DENSO VS-087, each with an ATI Axia80-M8 F/T sensor, a Robotiq 2F-85 gripper, and an Intel Realsense D405 as wrist camera, as well as a Realsense 435i as front camera and a Realsense 455 as top camera (shown in Fig. \ref{Fig: system}). The policy was trained on a server with 4 NVIDIA RTX 6000 Ada GPUs. The policy inference, phase prediction, and impedance controller were running on the same computer with an NVIDIA 4090, an Intel i9-14900KF CPU, and 64 GB of RAM, running Ubuntu 22.04 with ROS2.

\subsection{Datasets Collection and Training}

\del{To manage the complexity of the DOM, we decoupled the long-horizon process into defined phases: 5 phases for the hanging task and 3 phases for the taking-off task.
This ensures that the result of each sub-task is relatively controlled.}
\rev{We decompose the long-horizon process into 5 phases for hanging and 3 for taking off, ensuring each sub-task produces a controlled outcome.}
Data collection was performed using the haptic teleoperation system proposed in Sec. \ref{sec:method}. Although inference was limited to 15 Hz, the datasets were recorded at 30 Hz to collect as much information as possible, including 100 nominal episodes for both tasks, totaling 0.85 hours for hanging and 0.71 hours for taking off. During collection, the operator manually labeled phase transitions via the Quest controller buttons.
Additionally, we specifically collected 30 episodes \rev{for each} failure-recovery scenario (Phases 8, 9, and 10), lasting 2.3, 1.4, and 5.2 minutes, respectively. 
\rev{Both the policy and the phase predictor are trained on this dataset.}

The phase-conditioned ACT policy was trained on 640$\times$480 RGB images with a chunk size of $N_h = 50$. The architecture consists of four encoder layers and two decoder layers with a hidden dimension of $d_\text{model}=512$. Training was conducted for \del{500,000 steps}\rev{60 epochs} with a batch size of 8 on 4 GPUs, using a fixed learning rate of $1\times10^{-5}$.

\subsection{Experiment Setup}

The system is designed to be capable of autonomously executing two reversible tasks-hanging and taking off-by dynamically conditioning on the estimated phase. We visualize the nominal execution flows in Fig. \ref{Fig: exp-Nominal_tasks}. \del{Crucially, to} \rev{To} validate the closed-loop robustness, we evaluate three distinct failure recovery mechanisms (Fig. \ref{Fig: exp-failure phase8}, and \ref{Fig: exp-failure phase9/10}). These recovery phases are not arbitrary but are derived from the most frequent kinematic failures observed in garment manipulation.

\textbf{Phase 8 (Insertion Snagging):} During the transition from approach (Phase 0) to insertion (Phase 1), the garment collar frequently snags on the hanger tip due to local misalignment. This contact produces a significant spike in external force, as demonstrated by summing the force vectors from the dual arms (Fig. \ref{Fig: exp-failure phase8})\cite{Kosuge}. We assume the phase predictor model can detect this signature, given the transition, the policy triggers a local adjustment (shaking) strategy to disengage the snag and insert the hanger, transitioning to Phase 1 upon completion.

\textbf{Phase 9 (Gross Misalignment):} Due to extreme initial grasping position errors, the garment may completely miss the hanger (landing either in front of or behind it), as shown in Fig. \ref{Fig: exp-failure phase9/10}. In this scenario, local adjustment is insufficient. The policy identifies this state and executes a global reset strategy: returning the garment to a neutral position and restarting the process from Phase 0.

\textbf{Phase 10 (Mid-air Sliding):} During the takeoff task, particularly when grasping near the collar (Phase 6), the t-shirt may prematurely slide off the hanger due to gravity and low friction. This triggers a transition to Phase 10, which executes a dynamic mid-air re-grasp to recover control of the falling garment, as shown in Fig. \ref{Fig: exp-failure phase9/10}.

\rev{
\subsection{Experiment Results}
}

\subsubsection{\rev{Successful Rate on Overall Tasks}
}

\begin{table*}[ht]
    \centering
    \caption{Evaluation of Task Success and Recovery Robustness (N=100 trails per task)}
    \label{tab: success_rate}
    \begin{tabular}{lcccccc} 
        \toprule
        \textbf{Task} & \textbf{Total Trails} & \textbf{Natural Success} & \textbf{Failure Occurrences} & \textbf{Detection Rate} &\textbf{Recovery Success} &\textbf{Final Success Rate} \\
        \midrule
        Hanging  & 100 & 56 & 44 & 40/44 (90.91\%) & 31/40 (78.9\%) & 87\% \\
        Taking off  & 100 & 88 & 12 & 8/12 (66.67\%) & 4/8 (50\%) & 92\% \\
        \bottomrule
    \end{tabular}
\end{table*}

We evaluated our condition policy in two modes to quantify the benefit of the recovery mechanism: (1) Open-loop baseline, where any failure state is counted as terminal; and (2) Closed-loop system, where recovery phases are active. We conducted 100 test episodes for each task in a relatively fixed scenario similar to that of the recorded datasets. The results are summarized in Table \ref{tab: success_rate}. Without the recovery mechanism, the success rates for hanging and taking off were limited to 56\% and 88\%, respectively. However, our closed-loop system successfully recovered from 40 of 44 contact failures during hanging and from 4 of 8 during takeoff, thereby significantly improving the success rates to 87\% and 92\%, respectively.

In this evaluation, each trial begins from the taking-off task and proceeds directly to hanging without resetting the initial configuration.
Consequently, the nominal success rate of hanging is strongly influenced by the outcome of the preceding taking-off task, which may leave the garment in suboptimal configurations, leading to nominal failures as shown in Table~\ref{tab: success_rate}. These configurations are specific to the grasping points on the garment and to the initial collar states.
The proposed mechanism improves robustness despite this variability: although the robot maintains the original grasp during a retry following the recovery phase, the recovery maneuver itself (e.g., shaking or retreating) induces stochastic variations in the collar's deformation.
This strategy effectively provides a form of implicit exploration, preventing the policy from getting stuck in repetitive failure loops.
Note that the lower recovery rate in Phase~10 (Takeoff) is mainly attributable to grasp failures arising from the system’s limited spatial perception---a limitation we discuss further in Sec.~\ref{sec:conclusion}.

\subsubsection{\rev{Ablation Study on Execution-Time Robustness}
}
\begin{table}[ht]
    \centering
    \caption{\rev{Ablation Study on Execution-time Robustness. }}
    \label{tab: ablation}
    \begin{tabular}{@{}l l c c c@{}}
        \toprule
        & \textbf{Model}  & \textbf{Nominal} & \textbf{Misalign.} & \textbf{Snagging} \\
        \midrule
        A & ACT            
            & 8/20\,(40\%)  & 0/10\,(0\%)   & 1/10\,(10\%) \\
        B & ACT-R           
            & 16/20\,(80\%) & 1/10\,(10\%)  & 5/10\,(50\%) \\
        C & ACT-MR          
            & 9/20\,(45\%)  & 0/10\,(0\%)   & 6/10\,(60\%) \\
        D & ACT-MR+Phase\textsubscript{Tok}       
            & 12/20\,(60\%) & 0/10\,(0\%)   & 2/10\,(20\%) \\
        E & ACT-MR+Phase\textsubscript{FiLM}       
            & \textbf{18/20\,(90\%)} & \textbf{8/10\,(80\%)} 
            & \textbf{10/10\,(100\%)} \\
        \bottomrule
        \multicolumn{5}{@{}p{0.95\columnwidth}@{}}{\footnotesize 
        \textit{R}: with recovery demonstrations; 
        \textit{MR}: multi-task (hanging + taking-off) with recovery; 
        \textit{Tok}: phase injected as an input token; 
        \textit{FiLM}: phase injected via token and FiLM modulation (Ours).
        \textit{Nominal}: 20 undisturbed hanging trials. 
        \textit{Misalignment}: 10 trials with manual displacement (3--5\,cm). 
        \textit{Snagging}: 10 trials with manually induced collar snagging. 
        All trials start from a consistent robot configuration 
        with randomized grasping points ($\pm$1--2\,cm near the collar).
    }
    \end{tabular}
\end{table}


To isolate the contribution of each component, we evaluate five model variants on the hanging task alone,
as summarized in Table~\ref{tab: ablation}.
Unlike Table~\ref{tab: success_rate}, where the initial garment state is determined by the preceding taking-off task and thus highly variable, here we start each trial with a randomized but bounded grasping point ($\pm$1--2\,cm near the collar), ensuring that the nominal condition does not inherently trigger failures.
Perturbations (misalignment and snagging) are then deliberately introduced during execution to test recovery capability.

\textbf{Recovery data is beneficial but insufficient.}
Comparing A and B, incorporating recovery demonstrations (Dagger~\cite{Dagger}) substantially improves nominal performance (40\%\,$\to$\,80\%) and enables partial snagging recovery. 
However, Model~B fails almost entirely due to misalignment: during Phase~1, the policy exhibits oscillatory motion near the decision boundary between insertion and retreat, a direct manifestation of state aliasing, in which visually similar observations map to contradictory actions.

\textbf{Multi-task data amplifies aliasing without explicit conditioning.} Model~C, trained on both hanging and taking-off data, degrades nominal performance below even the single-task baseline~A (45\% vs.\ 40\%). 
Adding a contradictory task introduces new aliased states, causing the policy to freeze when moderate contact forces are encountered. 
This result highlights that naively scaling data with opposing action trajectories is harmful without a mechanism to disambiguate them.

\textbf{Token-level conditioning shows limited effectiveness.}
Model~D adds the phase label as an input token. 
While nominal performance partially recovers (60\%), snagging recovery drops sharply (20\%), misalignment remains unchanged,
indicating that the phase signal is diluted among the 1200+ visual tokens during self-attention and therefore has limited influence on the encoder's feature extraction.

\textbf{FiLM modulation and closed-loop mechanism enable robust phase-specific behavior.}
Model~E applies the phase condition via FiLM layers that directly modulate the encoder. 
This yields consistent improvements across all conditions (90\%, 80\%, 100\%). 
We note that 5 of the 10 snagging recoveries were executed via Phase~9 (global reset) rather than the intended Phase~8 (local shaking). 
As well as the two nominal failures were triggered by incorrect phase predictions, indicating that overall system performance is bounded by the phase predictor's accuracy---a direction for future improvement.


\subsubsection{\rev{t-SNE Visualization on Phase Modulating}
}
\label{sec: t-sne}
\begin{figure}[t]
    \centering
    \subfigure[Phase\textsubscript{Tok} (Model D)]{
        \begin{minipage}[t]{0.46\linewidth}
            \centering
            \includegraphics[width=\linewidth]{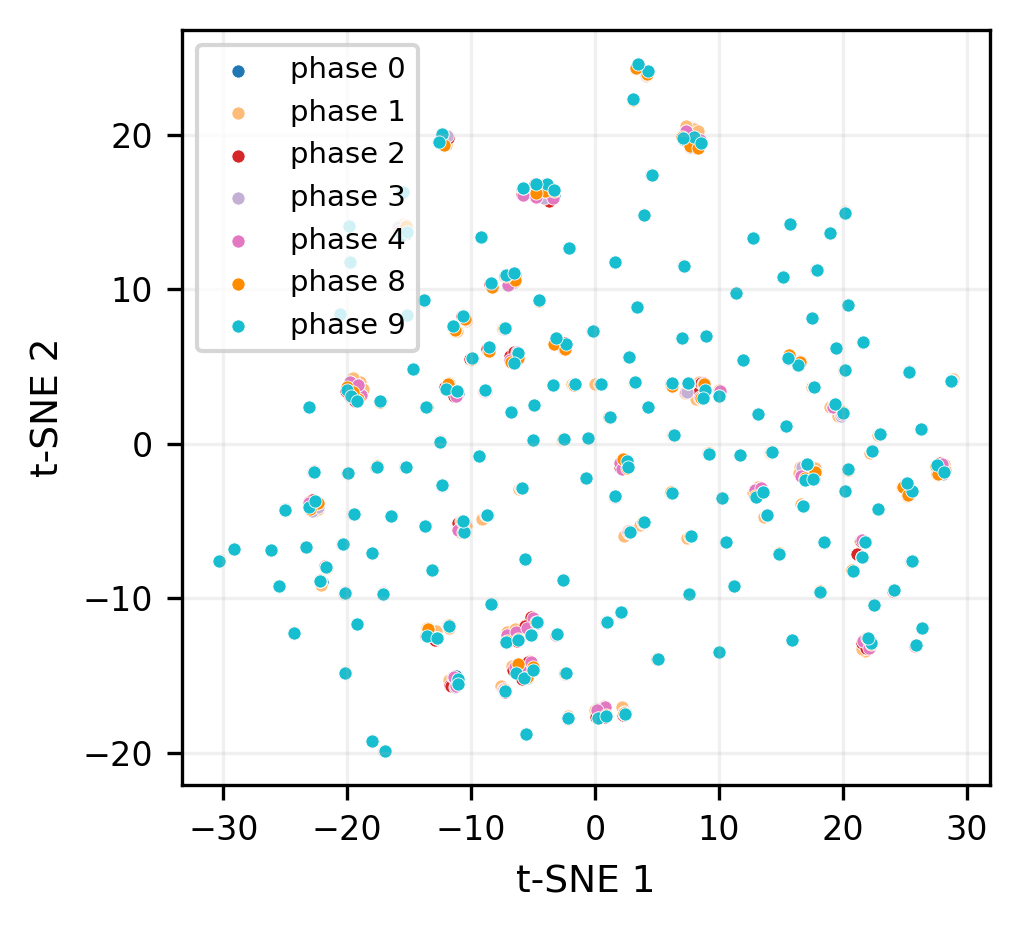}
        \end{minipage}
    }
    \subfigure[Phase\textsubscript{FiLM} (Model E, Ours)]{
        \begin{minipage}[t]{0.46\linewidth}
            \centering
            \includegraphics[width=\linewidth]{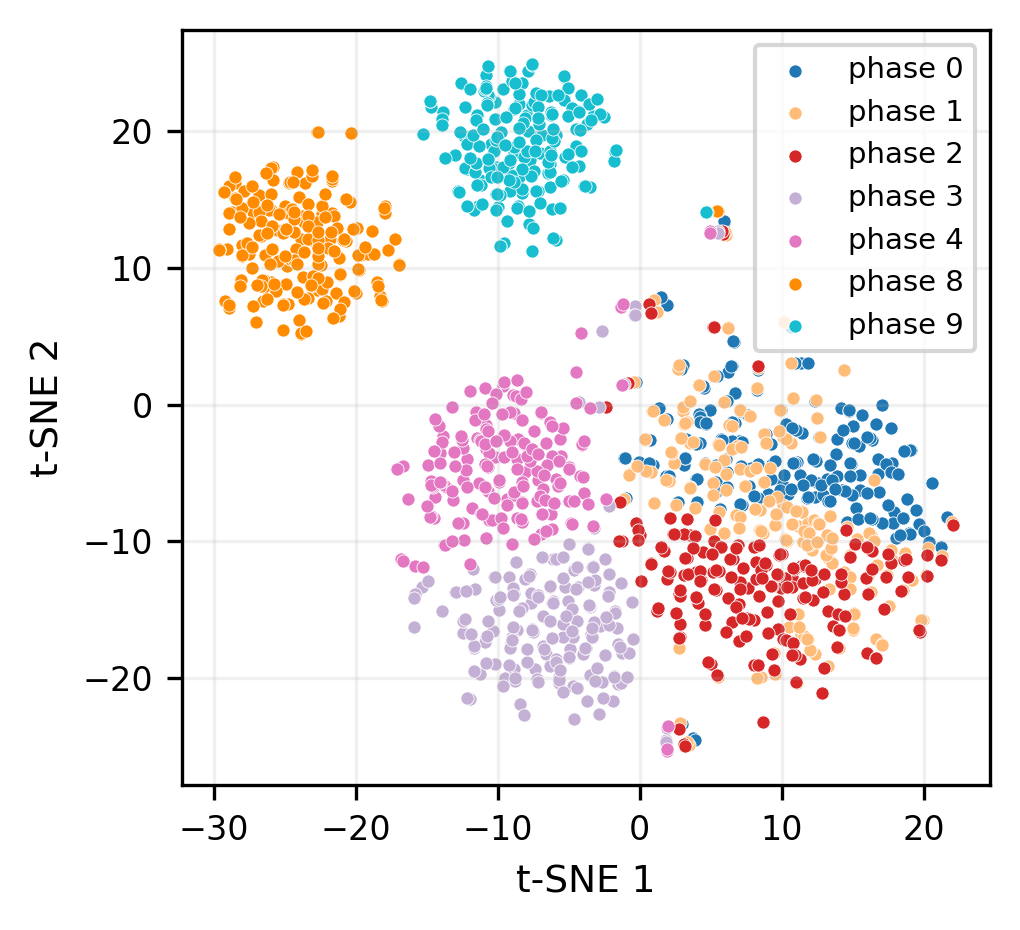}
        \end{minipage}
    }
    \caption{\rev{t-SNE visualization of ACT encoder outputs under varying phase conditions. The same 200 observations, randomly sampled from the hanging-task dataset Phase 1 subgroup, are each forwarded through the encoder with all seven phase labels (0--4, 8, 9), producing 1400 embedding vectors per model. Colors denote the phase condition.}}
    \label{Fig: t-sne analyzing}
\end{figure}

\rev{
To examine how phase conditioning modulates the encoder's internal representations, we apply t-distributed Stochastic Neighbor Embedding (t-SNE) to visualize the encoder output embeddings under different phase conditions, as shown in Fig.~\ref{Fig: t-sne analyzing}. Both subfigures use the identical set of input observations. Only the phase condition varies.}

\rev{
In Model~D (Fig.~\ref{Fig: t-sne analyzing}a), where the phase is injected as an input token, the embeddings for different phases remain heavily entangled. The encoder’s feature extraction is dominated by the visual input, and the phase token is ignored during self-
attention, which explains Model~D's poor discriminability between nominal and recovery behaviors in Table~\ref{tab: ablation}.}

\rev{
In contrast, Model~E (Fig.~\ref{Fig: t-sne analyzing}b) exhibits well-separated clusters for each phase. Since FiLM directly modulates the layer normalization and intermediate activations of every encoder layer, the same observation produces fundamentally different feature representations depending on the phase condition. Notably, the recovery phases (8, 9) form tight, isolated clusters far from the nominal phases, confirming that FiLM enables the encoder to switch between distinct perceptual modes rather than merely appending a condition signal that can be diluted by visual features.
}

\section{Conclusion and Discussion}\label{sec:conclusion}

\del{
This paper proposed an end-to-end solution based on imitation learning. By introducing a phase-conditioned network architecture that decomposes a long-horizon task into several subtasks, the model can focus on a specific task and reduce the difficulty of task execution. Then, a state-control network is proposed as a finite-state machine controller that introduces a failure-recovery mechanism to improve system robustness. Based on this mechanism, our method not only works for decomposing a long-horizon task into several subtasks but also for two tasks that are opposite, hanging and taking off, without confusion. Although current task success is highly dependent on dataset quality, with carefully collected datasets and failure state detection, this architecture has enormous potential to solve more complex tasks and scale.}
\rev{This paper presented a phase-conditioned imitation learning framework for robust deformable object manipulation.
By injecting phase conditions via FiLM into the ACT encoder, a single unified policy produces deterministic, phase-specific behaviors without confusion from state aliasing, even when trained on multi-task data with contradictory trajectories.
Building on this, we designed a closed-loop system in which a force-aware phase predictor monitors task progress and autonomously triggers failure recovery. Ablation studies confirmed that FiLM-level modulation is essential, rather than simply injecting the condition as an input token.
And t-SNE analysis provided visual evidence of phase-separated feature representations.
The closed-loop recovery mechanism improved the hanging success rate from 56\% to 87\% through autonomous error detection and recovery.
}

\rev{\textbf{Dynamic operation.} The current system operates at 15\,Hz policy inference with smooth, deliberate motions. We investigated higher-speed execution by selecting every 3rd and 5th action from the output chunk.
This reduced trajectory smoothness and action precision; in the taking-off task, the total execution time paradoxically exceeded the nominal speed, as the policy skipped fine-grained manipulation steps.
We attribute this to two factors: (1) subsampling the action chunk bypasses the temporal ensembler's smoothing 
mechanism, producing jerky motion; and (2) faster execution drives the manipulated object into out-of-distribution states not covered by the training data, causing policy confusion. Addressing this would likely require either collecting demonstrations at varying speeds or incorporating online reinforcement learning to adapt to dynamic conditions.}

\rev{\textbf{Generalization.} The robustness addressed in this work is execution-time robustness. Generalization-level robustness, such as adapting to different garment colors, materials, or shapes, is a separate challenge that this framework does not target. The current system relies on RGB inputs and is trained on a single T-shirt type; changing the garment's visual appearance significantly degrades performance without retraining. However, the framework itself---phase-conditioned policy, phase predictor, and closed-loop recovery---is object-agnostic: extending to new garments requires collecting phase-labeled demonstrations and retraining, without modifying the architecture.} Introducing explicit geometric features 
via point clouds or depth maps~\cite{Kai} could reduce the dependency on RGB appearance and improve generalization robustness, which we leave for future work.

\section{Acknowledgments}
This work was supported in part by the Innovation and Technology Commission of the HKSAR Government under the InnoHK initiative, and in part by the JST Moonshot R\&D under
Grant JPMJMS2034-18.
The author would also like to acknowledge LeRobot, an outstanding open-source framework \cite{Lerobot}, for facilitating rapid implementation and iteration of the imitation learning policies presented in this work. Large language models were used for grammar checking and sentence refinement during the preparation of this manuscript.


\bibliographystyle{IEEEtran}
\bibliography{cite}

\begin{IEEEbiography}
[{\includegraphics[width=1in,height=1.25in,clip,keepaspectratio]{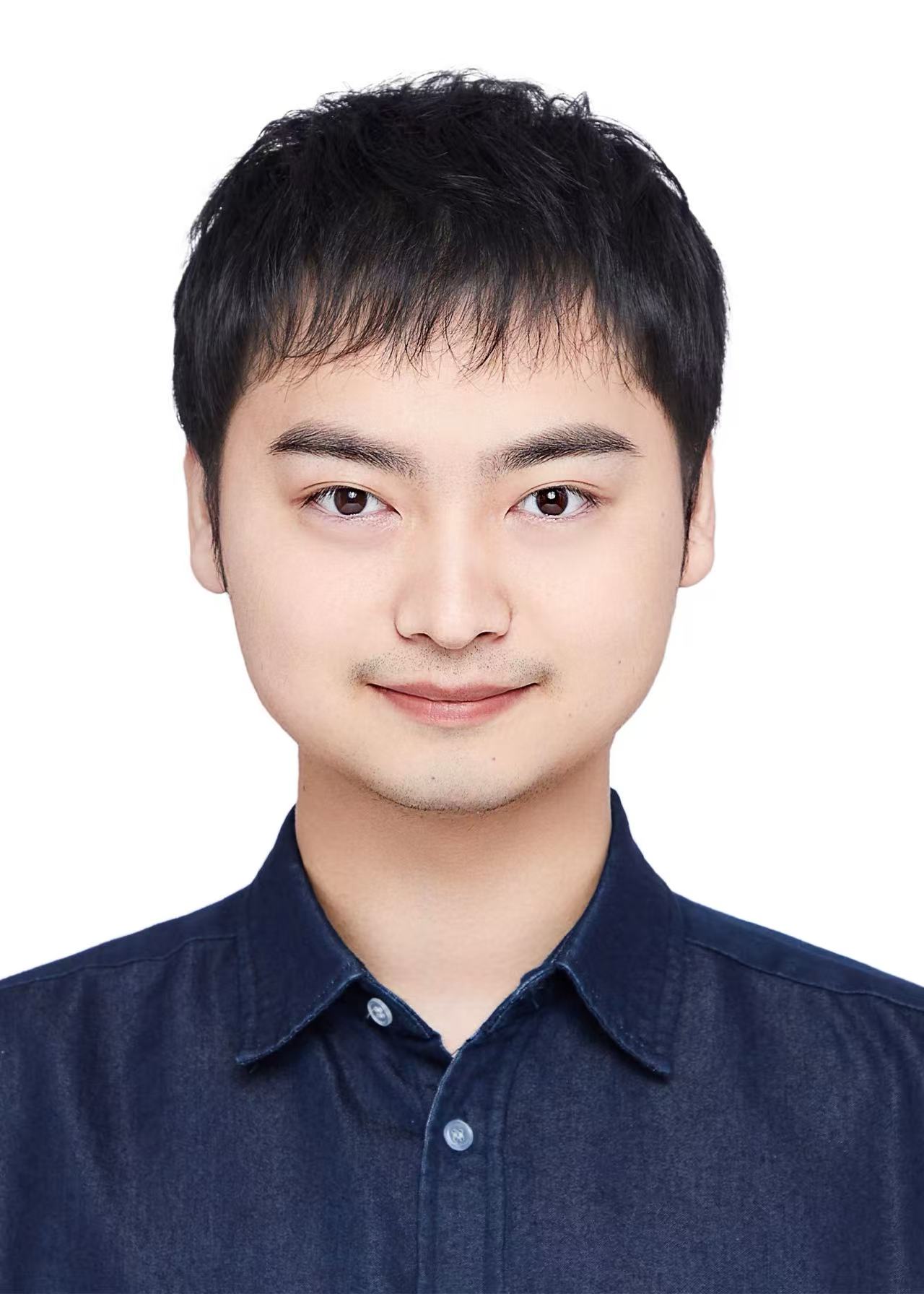}}]{Dayuan Chen} (Graduate Student Member, IEEE) received the B.E. in mechanical design, manufacturing, and automation from Dalian Minzu University, China, in 2016, and the M.E. in mechanical engineering from Shenzhen University, China, in 2019. He is currently pursuing his Ph.D. in robotics at Tohoku University, Sendai, Japan. He was an intern at Milebot Robotics Co., Ltd. from 2017 to 2019, where he focused on exoskeleton robots. Further, he expanded his research scope as a Research Assistant with Harbin Institute of Technology, Shenzhen, China, delving into construction robots and deformable object manipulations. His current research focuses on garment manipulation using robotic arms, leveraging force-aware imitation learning.
\end{IEEEbiography}

\begin{IEEEbiography}
[{\includegraphics[width=1in,height=1.25in,clip,keepaspectratio]{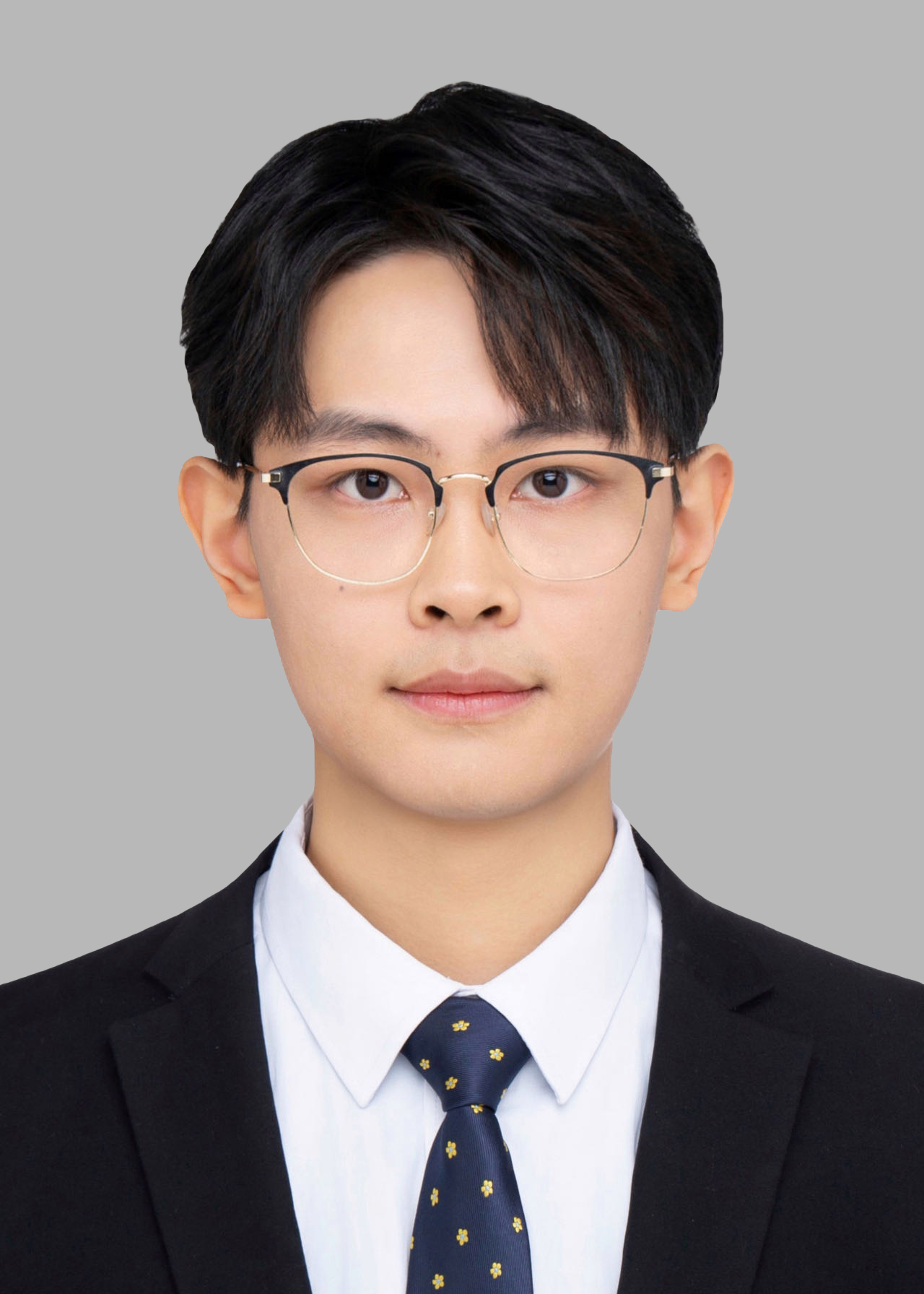}}]{Kai Tang} (Graduate Student Member, IEEE) received his B.Eng. degree in Process Equipment and Control Engineering from South China University of Technology in 2020, and M.Sc. degree with Distinction in Control Systems from Imperial College London in 2021. He is currently pursuing his Ph.D. degree in robotics at the JC STEM Lab of Robotics for Soft Materials, The University of Hong Kong. From 2022 to 2025, he worked with the Centre for Transformative Garment Production, Hong Kong SAR, which was in collaboration with Tohoku University, Japan. His research focuses on robot learning and control for fabric manipulation and fixture-free automated sewing.
\end{IEEEbiography}

\begin{IEEEbiography}
[{\includegraphics[width=1in,height=1.25in,clip,keepaspectratio]{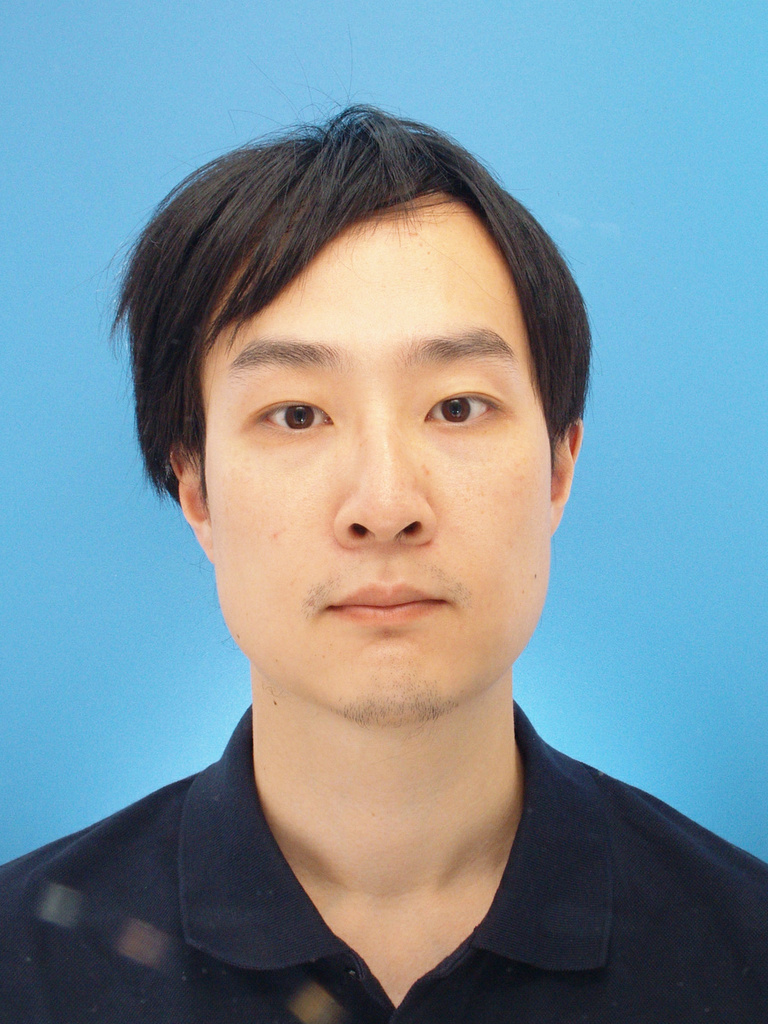}}]{Yukuan Zhang} (Member, IEEE) received the B.E. degree in Information Science and Technology from the University of Science and Technology of China in 2019, the M.E. and Ph.D. degree in Mechanical Engineering from Tohoku University, Sendai, Japan, in 2021 and 2025, respectively. He is currently a Reinforcement Learning Engineer at Anker Innovations Co., Ltd. His research interests include simulation systems, reinforcement learning, learning control systems, and reinforcement post-training based on simulation and world action models.
\end{IEEEbiography}

\begin{IEEEbiography}[{\includegraphics[width=1in,height=1.25in,clip,keepaspectratio]{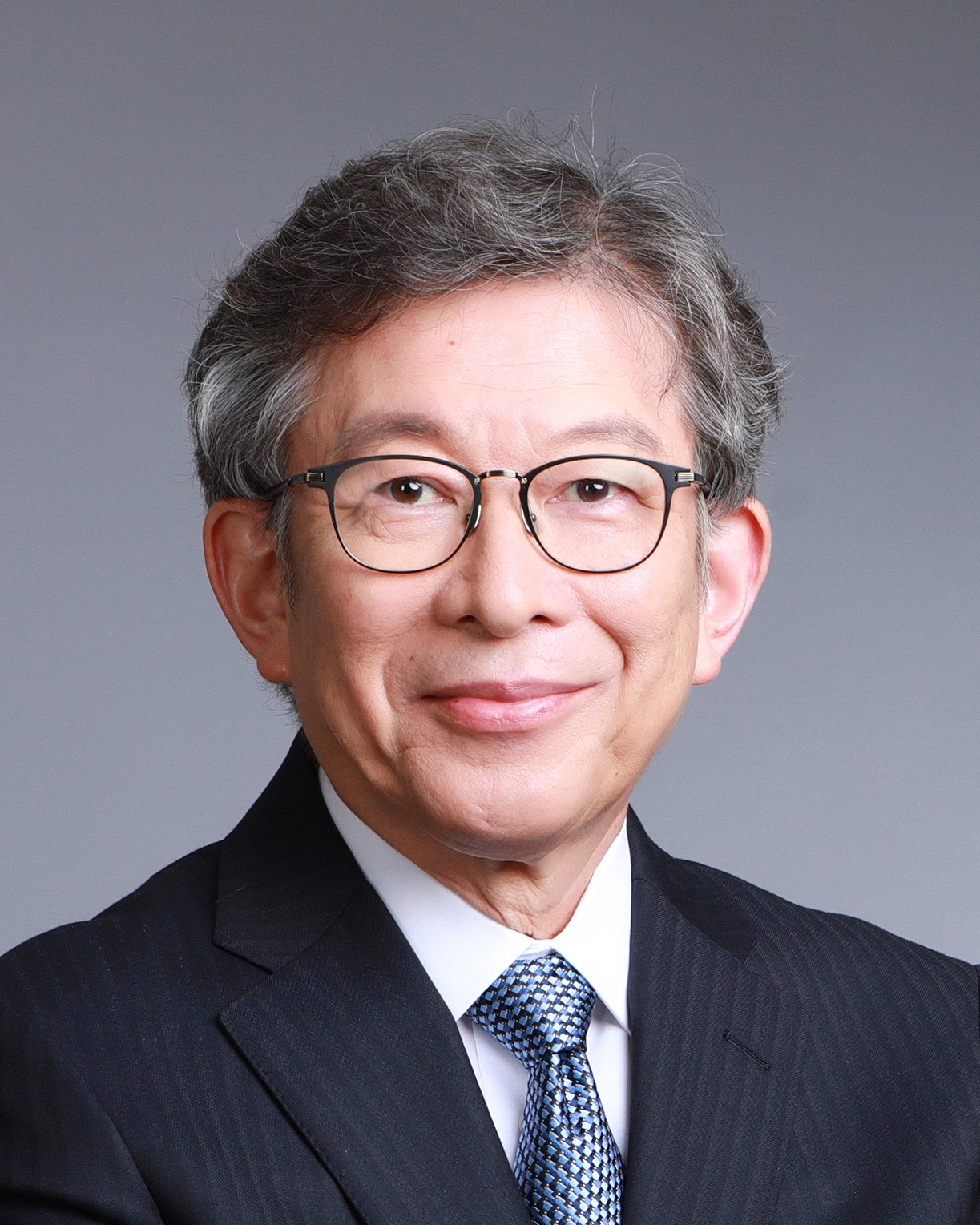}}]{Kazuhiro Kosuge} (Life Fellow, IEEE) received the B.S., M.S., and Ph.D. in control engineering from the Tokyo Institute of Technology, in 1978, 1980, and 1988 respectively. After having served as a R\&D Staff of the Production Engineering Department, Nippon Denso Company, Ltd., a Research Associate at Tokyo Institute of Technology, and an Associate Professor at Nagoya University, he joined Tohoku University as Professor in 1995 and served as Distinguished Professor from 2018 to March 2021. He is currently a Deputy Managing Director of the Centre for Transformative Garment Production, Hong Kong SAR, and the Director of the JC STEM Lab of Robotics for Soft Materials, Department of Electrical and Electronic Engineering, the University of Hong Kong, Hong Kong SAR.
    
He received the Medal of Honor, Medal with Purple Ribbon, from the Government of Japan in 2018 - a national honor in recognition of his prominent contributions to academic and industrial advancements. He also received IEEE RAS George Saridis Leadership Award in Robotics and Automation in 2021 for his exceptional vision of innovative research and outstanding leadership in the robotics and automation community through technical activity management. He is an IEEE Fellow, JSME Fellow, SICE Fellow, RSJ Fellow, JSAE Fellow and a member of the Engineering Academy of Japan. He was the President of the IEEE Robotics and Automation Society, from 2010 to 2011, the IEEE Division X Director, from 2015 to 2016, and the IEEE Vice President for Technical Activities for 2020.
\end{IEEEbiography}

\begin{IEEEbiography}
[{\includegraphics[width=1in,height=1.25in,clip,keepaspectratio]{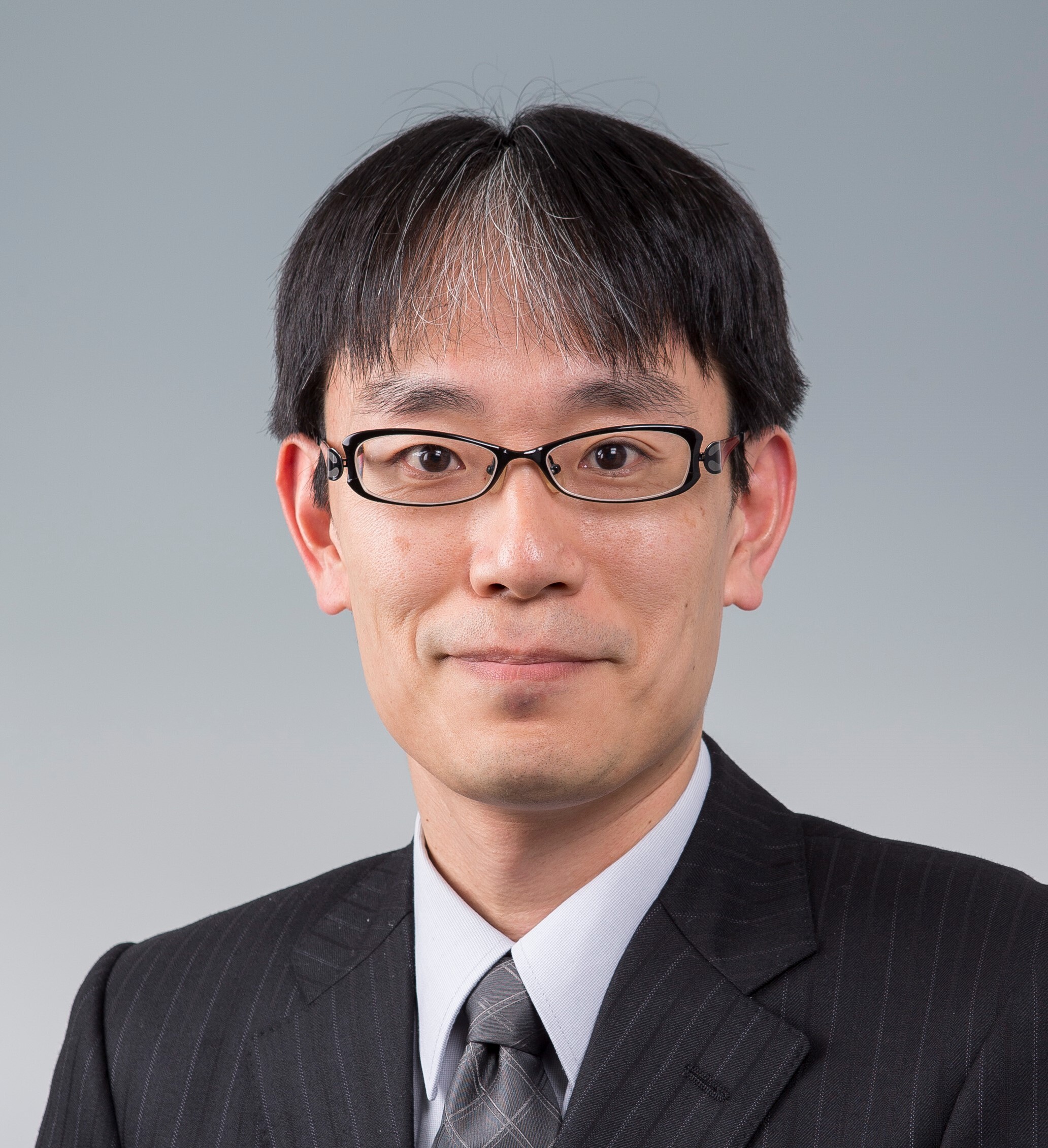}}]{Yasuhisa Hirata} (Senior Member, IEEE) received the B.E., M.E., and Ph.D. degrees in mechanical engineering from Tohoku University in 1998, 2000, and 2004, respectively. Currently, he holds the position of a Professor with the Department of Robotics, Tohoku University, Sendai. From 2020 to 2025, he served as the Project Manager of the Moonshot Research and Development Program, JST. His diverse research endeavors focus on assistive robotics, human–robot interaction, cooperative robotics, and manufacturing robotics. He is actively engaged in professional roles, having contributed as an AdCom Member of the IEEE Robotics and Automation Society (RAS) and fulfilling the position of the Chair for the Health and Medical Robotics Cluster.
\end{IEEEbiography}

\vfill

\end{document}